\documentclass{llncs}
\usepackage{amsmath, amssymb,bm}
\usepackage{graphicx}
\usepackage{subfigure}
\usepackage{xcolor}
\usepackage{booktabs}

\usepackage{comment}
\usepackage{bbm}
\usepackage{multirow}
\usepackage{threeparttable}
\usepackage{gensymb}
\usepackage{bbm}
\usepackage{tabularx}
\usepackage{tabu}
\usepackage{array}
\usepackage[pagebackref=true,breaklinks=true,letterpaper=true,colorlinks,bookmarks=false]{hyperref}
\newcommand{\afn}{\textsc{AFNet}}

\begin{document}
\title{Autofocus Layer for Semantic Segmentation}
\vspace{-3mm}
\author{Yao Qin\inst{1,2 \star}, %index{Qi, Dou}
Konstantinos Kamnitsas\inst{1,3\star}, %index{Hao, Chen}
Siddharth Ancha\inst{1,4}\thanks{Most of this work was performed while the authors were interns at MSR, Cambridge.}, %index{Yueming, Jin}
Jay Nanavati\inst{1}, %index{Huangjing, Lin}
\\ Garrison Cottrell\inst{2}, %index{Jing, Qin}
Antonio Criminisi\inst{1},
\and Aditya Nori\inst{1}} %index{Pheng-Ann, Heng}
\institute{Microsoft Research, Cambridge, UK \and University of California, San Diego, USA 
\and  Imperial College London, UK
\and Carnegie Mellon University, USA}
\maketitle
\vspace{-7mm}
\begin{abstract}
We propose the autofocus convolutional layer for semantic segmentation with the objective of enhancing the capabilities of neural networks for multi-scale processing.
Autofocus layers adaptively change the size of the effective receptive field 
based on the processed context to generate more powerful features.
This is achieved by parallelising multiple convolutional layers with different dilation rates, combined by an attention mechanism that learns to focus on the optimal scales driven by context. By sharing the weights of the parallel convolutions we make the network scale-invariant, with only a modest increase in the number of parameters.
The proposed autofocus layer can be easily integrated into existing networks to improve
a model's representational power. We evaluate our models on the challenging tasks of multi-organ segmentation in pelvic CT and brain tumor segmentation in MRI and achieve very promising performance.
\end{abstract}
\vspace{-5mm}
\section{Introduction}
\vspace{-3mm}
Semantic segmentation is a fundamental problem in medical image analysis. Automatic segmentation systems can improve clinical pipelines, facilitating quantitative assessment of pathology, treatment planning and monitoring of disease progression. They can also facilitate large-scale research studies, by extracting measurements from magnetic resonance images (MRI) or computational tomography (CT) scans of large populations in an efficient and reproducible manner.

For high performance, segmentation algorithms are required to use multi-scale context \cite{galleguillos2010context}, while still aiming for pixel-level accuracy. Multi-scale processing provides detailed cues, such as texture information of a structure, combined with contextual information, such as a structure's surroundings, which can facilitate decisions that are ambiguous when based only on local context. Note that such a mechanism is also part of the human visual system, via foveal and peripheral vision.

A large volume of research has sought algorithms for effective multi-scale processing. An overview of traditional approaches can be found in \cite{galleguillos2010context}. Contemporary segmentation systems are often powered by convolutional neural networks (CNNs). The various network architectures proposed to effectively capture image context can be broadly grouped into three categories. The first type creates an image pyramid at multiple scales. The image is down-sampled and processed at different resolutions. Farabet \textit{et al.} trained the same filters to perform on all such versions of an image to achieve scale invariance~\cite{farabet2013learning}. In contrast, DeepMedic \cite{Kamnitsas2017EfficientM3} proposed learning dedicated pathways for several scales, to enable 3D CNNs to extract more patterns from a larger context in a computationally efficient manner. The second type uses an encoder that gradually down-samples to capture more context, followed by a decoder that learns to upsample the segmentations, combining multi-scale context using skip connections \cite{long2015fully}. Later extensions include U-net \cite{ronneberger2015u}, which used a larger decoder to learn upsampling features instead of segmentations as in \cite{long2015fully}. Learning to upsample with a decoder, however, increases model complexity and computational requirements, when downsampling may not be even necessary. Finally, driven by this idea, \cite{chen2016deeplab,yu2015multi} proposed dilated convolutions to process greater context without ever downsampling the feature maps. Taking it further, DeepLab~\cite{chen2016deeplab} introduced the module Atrous Spatial Pyramid Pooling (\textsc{Aspp}), where dilated convolutions with varying rates are applied in parallel to capture multi-scale information. The activations from all scales are naively fused via summation or concatenation.

We propose the \textit{autofocus layer}, a novel module that enhances the multi-scale processing of CNNs by learning to select the `appropriate' scale for identifying different objects in an image.
Our work on autofocus shares similarities with \textsc{Aspp} in that we also use parallel dilated convolutional filters to capture both local and more global context.
The crucial difference is that instead of naively aggregating features from all scales, the autofocus layer adaptively chooses the optimal scale to focus on in a data-driven, learned manner. 
In particular, our autofocus module uses an attention mechanism \cite{bahdanau2014neural} to indicate the importance of each scale when processing different locations of an image (Fig.~\ref{resilient}). The computed attention maps, one per scale, serve as filters for the patterns extracted at that scale. Autofocus also enhances interpretability of a network as the attention maps reveal how it locally `zooms in or out' to segment different context. Compared to the use of attention in \cite{chen2016attention}, our solution is modular and independent of architecture.

We extensively evaluate and compare our method with strong baselines on two tasks: multi-organ segmentation in pelvic CT and brain tumor segmentation in MRI. We show that thanks to its adaptive nature, the autofocus layer copes well with biological variability in the two tasks, improving performance of a well-established model. Despite its simplicity, our system is competitive with more elaborate pipelines, demonstrating the potential of the autofocus mechanism. Additionally, autofocus can be easily incorporated into existing architectures by replacing a standard convolutional layer.

\vspace{-3mm}
\section{Method}
\vspace{-2mm}
\subsection{Dilated convolution}
As they are fundamental to our work, we first present the basics of dilated convolutions~\cite{chen2016deeplab,yu2015multi} while introducing notation.
The standard 3D dilated convolutional layer at depth $l$ with dilation rate $r$ can be represented as a mapping $\textbf{Conv}^{r}_l: {\textbf{F}_{l-1}} \rightarrow \textbf{F}^r_l$, 
where $\textbf{F}_{l-1} \in \mathbb{R}^{W' \times H' \times D' \times C'}$ and $\textbf{F}_l^r \in \mathbb{R}^{W \times H \times D \times C}$ are input and output tensors with $C$ channels (feature maps) of size ($W \times H \times D$). For neurons in $\textbf{F}_l^r$ , the size $\bm{\phi}^{\{x,y,z\}}_l \in \mathbb{N}^3$ of their receptive field on the input image can be controlled via $r$. 
For dilated convolution layers with kernel size $\bm{\theta}^{\{x,y,z\}}_l\in \mathbb{N}^3$, $\phi^{\{x,y,z\}}_l$ can be derived recursively as follows:
\vspace{-2mm}
\begin{equation}\label{rec}
\vspace{-2mm}
\bm{\phi}^{\{x,y,z\}}_l = \bm{\phi}^{\{x,y,z\}}_{l-1} +  r_l (\bm{\theta}^{\{x,y,z\}}_l-\mathbf{1}) \bm{\eta}_l^{\{x,y,z\}},
\end{equation}
Here $\bm{\eta}_l^{\{x,y,z\}} \in \mathbb{N}^3$ denotes the stride of the receptive field at layer $l$, which is a product of the strides of kernels in preceding layers. It can be observed from Eqn~(\ref{rec}) that greater context can be captured by increasing dilation $r_l$ but in less detail as the input signal is probed more sparsely. Thus greater $r_l$ leads to a `zoom out' behavior. Usually, the dilation rate $r$ is a hyperparameter that is manually set and fixed for each layer. Standard convolution is a special case when $r=1$. Below we describe the autofocus mechanism that adaptively chooses the optimal dilation rate for different areas of the input.

\begin{figure}[t]
\center
\includegraphics[width=\linewidth]{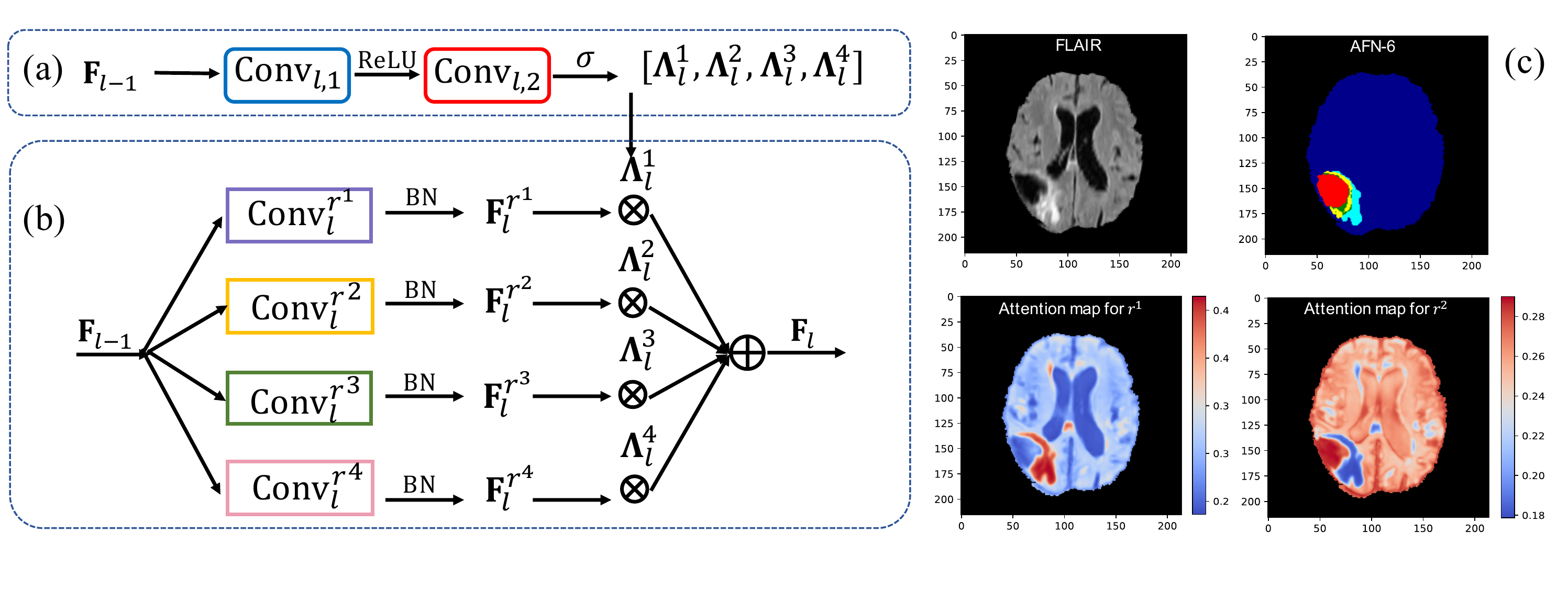}
\vspace{-10mm}
\caption{An autofocus convolutional layer with the number of candidate dilation rates $K=4$. (a) The attention model.
(b) A weighted summation of activations from parallel dilated convolutions. (c) An example of attention maps for a small ($r^1$) and larger ($r^2$) dilation rate. The first row is the input and the segmentation result of \textsc{Afn}-6 (described in Sec.~\ref{afn}). The second row shows how the module `zooms out' for more context when processing large or ambiguous structures.
}\label{resilient}
\vspace{-5mm}
\end{figure}
\vspace{-4mm}
\subsection{Autofocus convolutional layer}

Unambiguously classifying different objects in an image is likely to require different combinations of local and global information. For example, large structures may be better segmented by processing a large receptive field $\bm{\phi}_l$ at the expense of fine details, while small objects may require focusing on high resolution local information. Consequently, architectures that statically define multi-scale processing may be suboptimal. Our adaptive solution, the autofocus module, is summarized in Fig.~\ref{resilient} and formalized in the following.

Given activations of the previous layer $\textbf{F}_{l-1}$, we capture multi-scale information by processing it in parallel via $K$ convolutional layers with different dilation rates $r^k$. They produce $K$ tensors $\textbf{F}^{r^k}_l$ (Fig.~\ref{resilient}(b)), each set to have same number of channels $C$. They detect patterns at $K$ different scales which we merge in a data-driven manner by introducing a soft attention mechanism \cite{bahdanau2014neural}.

Within the module we construct a small attention network (Fig.~\ref{resilient}(a)) that processes $\textbf{F}_{l-1}$. In this work it consists of two convolutional layers. The first, $\textbf{Conv}_{l,1}$, applies 3$\times$3$\times$3 kernels, produces half the number of channels than those in $\textbf{F}_{l-1}$ (empirically chosen) and is followed by a ReLU activation function $f$. The second, $\textbf{Conv}_{l,2}$, applies 1$\times$1$\times$1 filters and produces a tensor with $K$ channels, one per scale. It is followed by an element-wise softmax $\sigma$ that normalizes the $K$ activations for each voxel to add up to one. Let this normalized output be $\mathbf{\Lambda}_l = [\mathbf{\Lambda}_l^{1}, \mathbf{\Lambda}_l^{2}, \cdots, \mathbf{\Lambda}_l^{K}] \in \mathbb{R}^{W \times H \times D \times K}$. Formally:
\vspace{-2mm}
\begin{equation}
\vspace{-1mm}
\mathbf{\Lambda}_l = \sigma(\textnormal{Conv}_{l, 2}(f(\textnormal{Conv}_{l,1}(\textbf{F}_{l-1}))))\\
%& = \sigma(\textbf{W}_2 \delta(\textbf{W}_1\textbf{F}_{l-1})),
\end{equation}

In the above, $\mathbf{\Lambda}_l^{k} \in \mathbb{R}^{W\times H \times D}$ is an attention map that corresponds to the $k$-th scale. For any specific spatial location (voxel), the corresponding $K$ values from the $K$ attention maps $\mathbf{\Lambda}_l^{k}$ can be interpreted as how much focus to put on each scale. Thus the final output of the autofocus layer is computed by fusing the outputs from the parallel dillated convolutions as follows:
\vspace{-3mm}
\begin{equation}
\textbf{F}_l = \sum_{k=1}^{K} \mathbf{\Lambda}_l^{k} \cdot \textbf{F}_l^{r^k}
\vspace{-3mm}
\end{equation}
where $\cdot$ is an element-wise multiplication. Note that the attention weights $\mathbf{\Lambda}_l^{k}$ are shared across all channels of tensor $\textbf{F}_l^{r^k}$ for scale $k$. Since the attention maps are predicted by a fully convolutional network, different attention is predicted for each voxel, driven by the image context for the optimal choice of scale (Fig.~\ref{resilient}(c)).

The increase in representational power offered by each autofocus layer naturally comes with computational requirements as the module is based in parallelism of $K$ dilated convolutional layers. Therefore an appropriate balance should be sought, which we investigate in Sec.~\ref{sec:evaluation} with very promising results. 

\vspace{-3mm}
\subsubsection{Scale invariance:}

The size of some anatomical structures such as bones and organs may vary, while the overall appearance is rather similar. For others, size may correlate with appearance. For instance, the texture of large developed tumors differs from early-stage small tumors. This suggests that scale invariance could be leveraged to regularize learning but must be done appropriately. We make the parallel filters in an autofocus layer share parameters. This makes the number of trainable parameters independent of $K$, with only the attention module adding parameters over a standard convolution. As a result, each parallel filter seeks patterns with similar appearance but of different sizes. Hence, the network is \textit{adaptively} scale-invariant -- the attention mechanism chooses the scale in a data-driven manner, unlike Farabet \emph{et al.}~\cite{farabet2013learning}, whose network learns shared filters between different scales but naively concatenates all their responses.

\begin{figure*}[t]
\center
\includegraphics[width=1\linewidth]{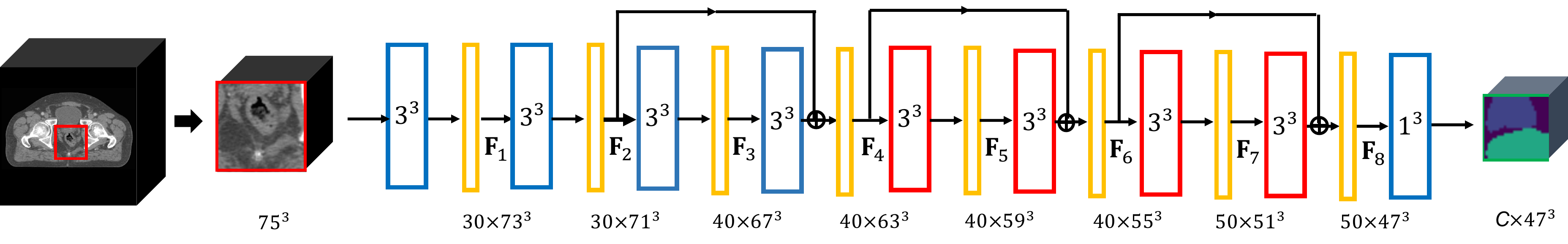}
\vspace{-2mm}
\caption{The \afn-4 model. Layers 1-2 are standard convolutions and 3-4 are dilated with rate 2. Layers 4-8 are autofocus layers, denoted with red. All layers except the classification layer use $3^3$ kernels. Yellow rectangles represent ReLU layers. Residual connections are used. Number and size of feature maps shown as (number $\times$ size). 
}\label{basic}
\vspace{-5mm}
\end{figure*}
\vspace{-4mm}
\subsection{Autofocus Neural Networks}
\label{afn}
\vspace{-1mm}

The proposed autofocus layer can be integrated into existing architectures to improve their multi-scale processing capabilities by replacing standard or dilated convolutions. To demonstrate this, we chose DeepMedic (\textsc{Dm}) ~\cite{Kamnitsas2017EfficientM3} with residual connections \cite{kamnitsas2016deepmedic} as a starting point. \textsc{Dm} uses different pathways with high and low resolution inputs for multi-scale processing. Instead, we keep only its high-resolution pathway and seek to empower it with our method. First, we enhance it with standard dilated convolutions with rate 2 in its last 6 hidden layers to enlarge its receptive field, arriving at the \textsc{Basic} model that serves as another baseline. We now define a family of \afn{}s by converting the last $n$ hidden layers of \textsc{Basic} to autofocus layers---denoted as ``\textsc{Afn}-$n$'', where $n \in \{1, \ldots , 6\}$. Fig.~\ref{basic} shows \afn-4. The proposed \afn s are trained end-to-end.

\vspace{-3mm}
\section{Evaluation}
\vspace{-3mm}
\label{sec:evaluation}
We extensively evaluate \afn{}s on the tasks of multi-organ and brain tumor segmentation. Specifically, on both tasks we perform: (1) a study where we successively add autofocus to more layers of the \textsc{Basic} network to explore its impact, and (2) comparison of \afn s with baselines. Finally, (3) we evaluate on the public benchmark BRATS'15 and show that our method competes with state-of-the-art pipelines regardless its simplicity, showing its potential.
\vspace{-5mm}
\subsubsection{Baselines:} We compare \afn{}s with the previously defined \textsc{Basic} model to show the contribution of autofocus layer over standard dilated convolutions. Similarly, we compare with DeepMedic~\cite{Kamnitsas2017EfficientM3}, denoted as \textsc{Dm}, to compare our adaptive multi-scale processing with the static multi-scale pathways. Finally, we place an \textsc{Aspp} module ~\cite{chen2016deeplab} on top of \textsc{Basic}, comparison of which against \textsc{Afn}-1 shows contribution of the attention mechanism. \textsc{Aspp-c} and \textsc{Aspp-s} represent fusion of \textsc{Aspp} activations via concatenation and summation respectively. Source codes and pretrained models in PyTorch framework are online available at: \url{https://github.com/yaq007/Autofocus-Layer}.

\vspace{-4mm}
\subsection{ADD and UW datasets of pelvic CT scans }
\label{subsec:ct_data}
\vspace{-2mm}
\textbf{Material:} We use two databases of pelvic CT scans, collected from patients diagnosed with prostate cancer in different clinical centers. The first, referred to as \textsc{Add}, contains 86 scans with varying number of 512x512 slices and 3mm inter-slice spacing. \textsc{Uw} consists of 34 scans of 512x512 slices with 1mm inter-slice spacing. Expert oncologists manually delineated in all images the following structures: prostate gland, seminal vesicles (SV), bladder, rectum, left femur and right femur. Each scan is normalized so that its intensities have zero mean and unit variance. We also re-sample \textsc{Uw} to the spacing of \textsc{Add}. To produce a stringent test of the models' generalization, we train them for this multi-class problem using the \textsc{Add} data and then evaluate them on \textsc{Uw} data.

\textbf{Configuration details:} \textsc{Basic}, \textsc{Aspp} and \textsc{Afn} models were trained with the ADAM optimizer for 300 epochs to minimize the soft dice loss~\cite{milletari2016v}. Each batch consists of 7 segments of size $75^3$. The learning rate starts at 0.001 and is reduced to 0.0001 after 200 epochs. We use dilation rates $2$, $6$, $10$ and $14$ ($K=4$) for both \textsc{Aspp} and the autofocus modules. It takes around 20 hours to train an \afn~ with 2 NVIDIA TITAN X GPUs. Performance of DeepMedic was obtained by training the public software~\cite{Kamnitsas2017EfficientM3} with default parameters, but without augmentation and by sampling each class equally, similar to other methods.

\vspace{-3mm}
\subsection{Brain tumor segmentation data (BRATS 2015) }
\label{subsec:brats_data}
\vspace{-1mm}
\textbf{Material:} The training database of BRATS'15~\cite{menze2015multimodal} consists of multi-modal MR scans of 274 cases, along with corresponding annotations of the tumors. We normalize each scan so that intensities belonging to the brain have zero mean and unit variance. For our ablation study, we train all models on the same 193 subjects and evaluate their performance on 54 subjects. The subsets were chosen randomly, including both high and low grade gliomas. Results on the remaining 23 cases aren't reported as they were used for configuration during development. Following standard protocol, we report performance for segmenting the \emph{whole} tumor, \emph{core} and \emph{enhancing} tumor. Finally, to compare with other methods, we train \afn-6 on all 274 images, segment the 110 test cases of BRATS'15 (no annotations publicly available) and submit predictions for online evaluation.

\textbf{Configuration details:} Settings are similar to Kamnitsas \emph{et al.}~\cite{Kamnitsas2017EfficientM3} for a fair comparison. For each method in Table~\ref{brats} we report the average of three runs with different seeds.

\begin{table*}[t]
    \centering
    \small
    \begin{tabular}{lcccccccccc}
        \multicolumn{11}{c}{}\\
        \hline
            & \textsc{Basic}        & \textsc{Aspp-s}   & \textsc{Aspp-c}    &  \textsc{Dm}  & \textsc{Afn}-1          & \textsc{Afn}-2          & \textsc{Afn}-3          & \textsc{Afn}-4          & \textsc{Afn}-5          & \textsc{Afn}-6          \\
        \hline
 \texttt{Prostate} & 50.94 &   55.83 &  58.67& 69.66&  63.36 &  75.43 &  \textbf{76.66}&   75.30 &  73.81 &  76.15 \\

       \texttt{Bladder} &   72.83  &  81.53& 80.43  &93.54 & 88.76  & 92.38  & 93.56 &  \textbf{94.49}  & 94.28  & 93.32  \\
    \texttt{Rectum}&    64.39 &   67.30&  67.04& 70.74 &  71.46 &  76.20 &  78.45&   \textbf{79.80} &  78.96 &  78.82 \\
\texttt{SV}&   53.97  & 50.11 & 59.37 & 56.75  & 61.68  & 65.03  & \textbf{65.12} &  64.83  & 60.87  & 63.24  \\

       \texttt{LFemur}&  91.60   &  94.13 & 93.81&94.68  & 93.24  & \textbf{95.18}  & 93.52 &  94.59  & 93.42  & 95.16  \\

       \texttt{RFemur}&  91.65  &  94.34 & 92.78 &94.63  & 91.93  & 94.38  & 91.61 &  94.60  & 94.16  & \textbf{95.75}  \\
 \hline
         Mean Dice &     70.90 &     74.98 & 75.35& 78.89&78.41 &  83.10 &  83.15&   \textbf{83.94} &  82.58 &  83.74 \\
\hline
    \end{tabular}
    \caption{Performance on multi-organ segmentation problem of baseline models and AFN on \textsc{Uw} database, after being trained on \textsc{Add}. Absolute dice scores are shown.}
    \label{prostate}
    \vspace{-9mm}
\end{table*}

\begin{table*}[t]
    \centering
    \small

    \begin{tabular}{lcccccccccc}
        \multicolumn{10}{c}{}\\
    	\hline
         & \textsc{Basic}        & \textsc{Aspp-s} & \textsc{Aspp-c}       & \textsc{Dm}  & \textsc{Afn}-1          & \textsc{Afn}-2          & \textsc{Afn}-3          & \textsc{Afn}-4          & \textsc{Afn}-5          & \textsc{Afn}-6          \\
        \hline

       \multirow{2}{0cm}{\texttt{Whole}} &  $87.90$ &   $87.83$ &  $87.90$&  88.93&$88.03$ &  $88.63$ &  $88.42$ &   $88.88$ &  $89.19$ &  $\mathbf{89.30}$ \\
       \ & $( 8.57)$ &    $( 8.36)$ & $( 8.08)$ & (7.05)& $( 8.29)$ &  $( 8.02)$ &  $( 9.28)$ &   $( 8.31)$ &  $(7.87)$ &  $\mathbf{( 8.00)}$\\
       \hline
       \multirow{2}{*}{\texttt{Core}}  & $72.61$  &  $74.08$ & $73.58$  &71.42& $73.93$  & $73.43$   & $73.79$ &  $73.91$  & $\mathbf{74.32}$  & $74.11$  \\
       \ & $( 26.39)$  &   $( 24.16)$ &  $( 24.57)$& (27.48)& $( 25.04)$ &  $( 24.90)$ &  $( 26.00)$ &   $( 25.49)$ &  $\mathbf{( 24.80)}$ &  $( 24.19)$\\
       \hline
       \multirow{2}{*}{\texttt{Enh}} &  $74.37$ &   $73.06$ &  $73.47$&$\mathbf{76.08}$&  $73.17$ &  $73.94$ &  $74.21$&   $74.08$ &  $74.48$ &  $75.62$ \\
       \  & $( 25.23)$  &   $( 26.93)$ &  $( 25.36)$ &$\mathbf{(25.12)}$&  $( 26.98)$ &  $( 25.83)$ &  $( 27.58)$ &   $( 25.70)$ &  $( 26.56)$ &  $( 25.02)$\\
    \hline
    \end{tabular}
    \caption{Ablation study on BRATS'15 training database via cross-validation on 54 random held-out cases. Dice scores shown in format mean(standard deviation).}
    \label{brats}
    \vspace{-8mm}
\end{table*}

\begin{table}[t]
\centering
\small
\label{tab:model_params}
\begin{tabu} to \textwidth {X[c]X[c]X[c]X[c]X[c]X[c]X[c]}
\hline
Models       & \textsc{Basic}             & \textsc{Aspp-s}        & \textsc{Aspp-c}& \textsc{Dm}  & \textsc{Afn}-1&\textsc{Afn}-6          \\
\hline
Params & 315,725   &967,330  &478,435 & 662,555 &349,904& 450,209 \\
\hline
\end{tabu}
\caption{Number of trainable parameters in convolutional kernels of different models.}
\vspace{-9mm}
\label{para}
\end{table}

\begin{table}[]
\centering
\label{tab:brats15_test}
\begin{tabular}{ccccccc}
\multicolumn{7}{c}{} \\
\hline
Models	& \textsc{Afn}-6	& peres1\textsuperscript{\textdagger}\textsuperscript{\textasteriskcentered}\cite{pereira2016brain} 	& bakas1\textsuperscript{\textdagger}\cite{bakas2015glistrboost}    	& kamnk1/\textsc{Dm}~\cite{Kamnitsas2017EfficientM3} & kayab1\textsuperscript{\textasteriskcentered}~\cite{kayalibay2017cnn}  & isenf1\textsuperscript{\textasteriskcentered}~\cite{isensee2017brain}     \\
Whole	& 84\%				& 83\%   							& 81\%		& 84\%  									& 85\%     & 85\%     \\
Core	& 69\%              & 72\%   							& 63\%		& 67\%  									& 72\%     & 74\%   \\

Enh		& 63\%              & 60\% 								& 58\%		& 63\%                                      & 61\%     & 64\%    \\
Cases	& 110/110      		& 53/110 		& 53/110	& 110/110       & 110/110 & 110/110 \\
\hline
\end{tabular}
\caption{Dice scores achieved by state-of-the-art methods on BRATS'15 test database. \textsuperscript{\textdagger} are semi-automatic. \textsuperscript{\textasteriskcentered} used CNN ensembles and more extensive augmentation.}
\vspace{-9mm}
\label{lead}
\end{table}

\vspace{-3mm}
\subsection{Results}
\vspace{-2mm}
\subsubsection{Ablation study:}
Results from the ablation study on the cervical CT database and the BRATS database are summarized in Table~\ref{prostate} and Table~\ref{brats} respectively. We observe the following: (a) Building \textsc{Afn}-1 by converting the last layer of \textsc{Basic} to autofocus improves performance, while (b) the gains surpass those by the popular \textsc{Aspp} for most classes of the tasks. It is important to note that \textsc{Aspp} adds multiple parallel convolutional layers without sharing weights between them. This incurs a large increase in the number of  parameters,
and is therefore partly the reason for improvements of \textsc{Aspp} over \textsc{Basic} (see Table~\ref{para}). (c) Converting more layers of the \textsc{Basic} baseline to autofocus layers tends to improve performance. An exception is \textsc{Afn}-4 vs. \textsc{Afn}-5/6 on the \textsc{Uw} dataset. We speculate that this is due to randomness in training and suboptimal optimization. (d) Empowering the high-resolution pathway of DeepMedic with adaptive autofocus quickly outperforms the gains from the static second pathway on pelvic scan and brain tumor segmentation except for the enhancing tumor. We speculate that gains are more profound in the former task due to the greater variation in the size of structures, where the adaptive nature of autofocus shines. Finally we note that by sharing weights across scales, \afn{}s have small number of trainable parameters, shown in Table~\ref{para}, which could enable rapid learning from little data, which is however left for future work. On the downside, the multiple scales on each autofocus layer increase memory and computation requirements.

\vspace{-4mm}
\subsubsection{Comparison with state-of-the-art on BRATS'15:}
Performance on test data of BRATS'15 obtained via the online evaluation platform is shown on Table~\ref{lead}, along with other top published methods. \textsc{Afn}-6 compares favorably to the semi-automatic methods that topped the BRATS'15 challenge~\cite{bakas2015glistrboost,pereira2016brain}, as well as DeepMedic with the second static lower-resolution pathway. Note that in \cite{pereira2016brain} high and low grade gliomas were separated by visual inspection and then passed to an appropriately specialized CNN, giving them an advantage over other methods. Our model is only surpassed by the pipelines of~\cite{kayalibay2017cnn} and~\cite{isensee2017brain}, who both used ensembles of CNNs with deep supervision and more aggressive data augmentation. The promising performance obtained by our simple method indicates the potential of the autofocus layer, which can be adopted in more elaborate systems.

\vspace{-4mm}
\section{Conclusion}
\vspace{-3mm}
We proposed an autofocus convolutional layer for segmentation of biomedical images. An autofocus layer can adapt the network's receptive field at different spatial locations in a data-driven manner. 
Our extensive evaluation of \afn{}s shows that they cope well with biological variability in different tasks and generalize well on both MR and CT images. We have shown that the autofocus convolutional layer can be integrated into existing network architectures to substantially increase their representational power with only a small increase in model parameters. In addition, the interpretability of autofocus layers can leverage understanding of deep learning systems. Investigating the potential of autofocus modules in regression problems would be interesting future work.
\vspace{-3mm}
\section*{Acknowledgments}
\vspace{-3mm}
G.W.C. and Y.Q. were partially supported by Guangzhou Science and Technology Planning Project (Grant No. 201704030051).

\vspace{-2mm}
\bibliographystyle{splncs03}

\end{document}